\documentclass[conference]{IEEEtran}
\IEEEoverridecommandlockouts
\newcommand{\comment}[1]{}
\usepackage{cite}
\usepackage{amsmath,amssymb,amsfonts}
\usepackage{algorithmic}
\usepackage{graphicx}
\usepackage{textcomp}
\usepackage{xcolor}
\usepackage{calrsfs}
\DeclareMathAlphabet{\pazocal}{OMS}{zplm}{m}{n}

\usepackage{color,soul}
\usepackage{multirow}
\usepackage{booktabs,array,dcolumn}

\usepackage{subfigure}
\newcolumntype{d}{D{.}{.}{2.3}}
\newcolumntype{C}{>{\centering}p}
\usepackage{caption} 
\usepackage{float}
\restylefloat{table}

\def\BibTeX{{\rm B\kern-.05em{\sc i\kern-.025em b}\kern-.08em
  T\kern-.1667em\lower.7ex\hbox{E}\kern-.125emX}}
\begin{document}

\title{Semantically Enhanced Models\\for Commonsense Knowledge Acquisition
}


%

\author{\IEEEauthorblockN{Ikhlas Alhussien\IEEEauthorrefmark{1}, Erik Cambria\IEEEauthorrefmark{1}, Zhang NengSheng\IEEEauthorrefmark{2}}
\IEEEauthorblockA{\IEEEauthorrefmark{1}School of Computer Science and Engineering\\
Nanyang Technological University, Singapore\\
Email: ikhl0001@e.ntu.edu.sg, cambria@ntu.edu.sg} \IEEEauthorblockA{\IEEEauthorrefmark{2}Singapore Institute of Manufacturing Technology (SIMTech)\\ 
Agency for Science, Technology and Research (A*STAR), Singapore\\
Email: nzhang@simtech.a-star.edu.sg}}

\maketitle

\begin{abstract}
Commonsense knowledge is paramount to enable intelligent systems. Typically, it is characterized as being implicit and ambiguous, hindering thereby the automation of its acquisition. To address these challenges, this paper presents semantically enhanced models to enable reasoning through resolving part of commonsense ambiguity. The proposed models enhance in a knowledge graph embedding framework for knowledge base completion. Experimental results show the effectiveness of the new semantic models in commonsense reasoning. 	
\end{abstract}
	\section{Introduction}
Intelligent systems need to acquire human-like knowledge in order to perform smart decision making. This type of knowledge which is often termed commonsense knowledge refers to the agreed-upon facts and information about everyday world that is assumed to be shared by everyone. \comment{Unlike encyclopedic knowledge, commonsense is concerned with abstract concepts and classes rather than named entities and class instances.} Despite its abundance, commonsense knowledge is often excommunicated or expressed in implicit manner\comment{rarely expressed explicitly in textual corpora}. 

At early stages, researchers relied on manual annotation by systems experts to formalize and codify commonsense assertions~\cite{lenat1995cyc, singh2002open}. With advancements in machine learning and information extraction techniques, researchers turned to automating commonsense knowledge acquisition (CSKA) through inferring this knowledge from textual resources via pattern matching~\cite{pasca2014queries, clark2009large, etzioni2004web,tandon2011deriving,camacq,camgec,camnt5}. 
Curated resources have the advantage of having high precision, however, they tend to lack sufficient coverage while text mining techniques produce huge knowledge collections at the cost of low precision, in addition to being limited to the knowledge that is expressed in explicit manner and which is amenable for data mining.

Reasoning approaches, on the other hand, attempt to automatically infer missing knowledge based on pre-existing knowledge. This direction of CSKA go beyond literal extraction of explicit knowledge to elicitation of implicit assertions. By representing a knowledge base as a graph, a family of techniques referred to as knowledge graph embedding (KGE) convert knowledge graph entities and relations into $k$-dimensional vectors
and perform reasoning over knowledge graph vector model. These methods deliver eminent performance in enriching encyclopedic knowledge bases, such as DBpedia~\cite{lehmann2015dbpedia} and Freebase~\cite{bollacker2008freebase}, with missing facts. 

Nevertheless, such performance is not observed when KGE models are applied to commonsense knowledge bases, mainly because commonsense is associated with abstract and generic concepts, rather than named entities, which are related through non-functional semantic relations. 
Improvements can be promoted through incorporating auxiliary information that carry semantic knowledge associated with concepts into KGE models. Previous work in this direction applied with factual knowledge utilized entities' description~\cite{zhong2015aligning,xie2016representation}, Wikipedia anchors~\cite{wang2014knowledge}, newspapers~\cite{han2016joint}, entities' original phrasal form~\cite{li2016commonsense}, etc. Some approaches adopted more sophisticated context definition, such as graph paths~\cite{lin2015modeling, guu2015traversing,toutanova2016compositional} and syntactic parsing of entities mention~\cite{toutanova2015representing}. 

Inspired by these models, we advise a set auxiliary semantic information that is tailored to improve commonsense reasoning and incorporate them into compositional models for learning commonsense KGEs. Embeddings of semantic resources are trained individually then fine-tuned in a joint framework to ensure their compatibility with each other and with knowledge graph structure.
Our goal is to expand existing commonsense knowledge bases by augmenting them with missing assertions. We use semantically enhanced embeddings to perform knowledge base completion (KBC), a technique that perform reasoning over existing knowledge in supervised manner to predict missing assertions by filling missing elements of a triple (e.g., $(?,r,t))$. 

	The rest of the paper is organized as follows: Section 2 reviews related work; Section 3 provides a formal definition of the problem; Section 4 describes the proposed model; Section 5 presents experimental results; finally, Section 6 offers concluding remarks.


	\section{Related Work}

Earliest efforts for commonsense knowledge acquisition have relied on manual annotation by system experts to formalize and codify valid assertions including Cyc~\cite{lenat1995cyc}, SUMO ontology~\cite{niles2001towards} and Open Mind Common Sense (OMCS)~\cite{singh2002open}. To increase the efficiency of manual knowledge gathering, researchers resorted to collective efforts through public platforms such as crowd-sourcing websites and games with purpose~\cite{von2006verbosity}. 


A shift towards large-scale commonsense knowledge acquisition leveraged textual resources via pattern matching to discover potential valid assertions, limiting their scope to explicit or subtly implicit commonsensical sentences. Some papers relied on handcrafted extraction patterns~\cite{pasca2014queries, clark2009large, etzioni2004web}, while others followed bootstrapping method of pattern generation and fact extraction~\cite{tandon2011deriving}. Despite the high recall and the expanded coverage of these methods, they usually suffer from low precision.


The next step of knowledge acquisition is knowledge completion which relies on pre-existing knowledge to learn regularities in order to infer missing assertions~\cite{mintz2009distant,nickel2011three,nickel2012factorizing}. Vector space models convert entities and relations of knowledge base into compact k-dimensional vectors, and uses those representations to predict which facts are missing. For example, IsaCore~\cite{cammds} generates analogical closure of ProBase~\cite{wu2012probase} by applying singular value decomposition (SVD) on the knowledge base matrix representation. Other methods rely on neural networks architecture to obtain and use vector space representations~\cite{bordes2011learning, bordes2013translating,dong2014knowledge}


Our work aligns closely with neural network like methods. We re-cast the problem of commonsense knowledge acquisition as knowledge base completion in which we rely on vector space representations to perform reasoning. Li et al.~\cite{li2016commonsense} devised two models, bilinear and deep neural network, to embed concepts and provide scores to arbitrary triple. They considered concepts as phrasal terms and learned their representations through word embeddings trained over a limited training set, i.e., using the sentences underlying training triples. In contrast, we rely on a border set of semantic information resources that we import into knowledge base representation learning model via joint objective function to combine their local relational structure with their richer global semantics. 

Similarly, Chen et al.~\cite{chen2016webbrain} presented an approach for harvesting commonsense knowledge that relies on joint learning model from web-scale data. The model learn vector representations of commonsensical words and relations jointly using large-scale web information extractions and general corpus co-occurrences. Given triples of form $(subject, predicate, object)$, the model learns words representations of subject and object by optimizing Word2vec CBOW objective on general corpus while simultaneously optimize for modeling the explicit relationships in triples. We distance ourself from this work by learning separate entity representations for semantic information and explicit relations then enforce them to be compatible rather than unifying them into a single representation. 	



Word embeddings are learned based on context inputs and typically capture semantic similarities between words~\cite{mikolov2013linguistic, pennington2014glove}. Based on the type and scope of the context, word embeddings can vary between capturing generic to specific semantic, syntactic or lexical similarities. Abundant generic word embeddings learned over open corpora are available online\footnote{http://github.com/3Top/word2vec-api}. 

Joint models for knowledge base completion incorporate word embeddings of entities and relations to better transfer semantic knowledge between entities. Several models propose customized word embeddings that address the link predication task by training them in constraint context such as description~\cite{zhong2015aligning,xie2016representation} or lexical context~\cite{toutanova2015representing}. Drawing on the same idea, Chen et al.~\cite{chen2015neural} learned word embeddings that reflect commonsense information about words by simultaneously training on generic contexts from open corpus and semantically significant contexts, specifically, word definitions and synonyms as well as lists and enumerations. Similarly, Numberbatch~\cite{speer2017conceptnet} utilizes ConceptNet semantic network to define the context of a word as all other words to which it is connected in the network. 
	\section{Problem Formulation}
We begin by introducing the notation used in this paper. A commonsense knowledge base is represented as a graph $\pazocal{G} = \{ \pazocal{C,R,T} \}$, where $\pazocal{C}$ is the set of concepts, $\pazocal{R}$ is the set of relations, and $\pazocal{T}$ is the set of triples. Each triple represents head and tail concepts connected through a semantic relation, e.g., (\textit{Victory}, {\tt Causes}, \textit{Celebration}) and is denoted as $(h,r,t)$ such that $h,t \in \pazocal{C}$ and $r \in \pazocal{R}$. 



\begin{figure}[h]\centering
	\advance\rightskip-.1cm \belowcaptionskip-.5cm
	\includegraphics[width=\linewidth]{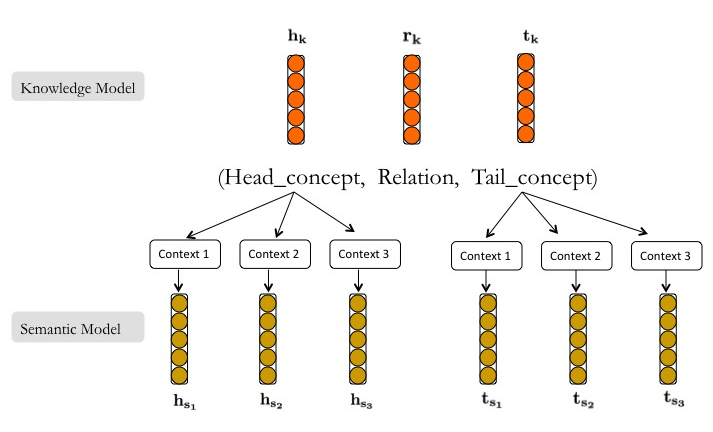} \caption{Model Architecture}
	\label{fig:model_architecture}
\end{figure}

Given a set of triples $\pazocal{T}$, our objective is to predict new commonsensical assertions that are not originally in the knowledge base by filling missing entries of incomplete triples of form $(h,t,?)$, $(?,r,t)$, or $(h,?,t)$. This task is termed knowledge base completion (KBC). To accomplish this, we aim to learn vector representations (alternatively embeddings) of concepts $\mathbf{h,t}$ and relations $\mathbf{r}$ in $\mathbb{R}^k$ \comment{utilizing structural and semantic information resources} such that we can measure a plausibility of a triple through a score function $f(\mathbf{h,r,t})$ over its embeddings. Our proposed framework consists of two models: knowledge representation and semantic representation (Fig.~\ref{fig:model_architecture}). \\


\textbf{Knowledge Representation Model:} Learn entities and relations representations directly from the explicit relationships within each triple via KGE method. Each concept $c$ and relation $r$ has a knowledge-based vector representation $\mathbf{c_k}$ and $\mathbf{r_k}$, respectively. \\


\textbf{Semantic Representation Model}: Learn entities representations from external information resources that retain some of concepts semantics, e.g., concept description, concept original phrase form, concept definitions, and many others. In this work, each concepts $c$ has a set of semantic descriptions $S_c$, such that $S_j$ is the $j^{th}$ class of semantic descriptions and $s_{i,c}$ is the $i^{th}$ semantic description of of concept $c$. Each entity has separate representation $\mathbf{c_{s_i}}$ for each semantic description $s_{i,c}$.
\hspace*{-7in}

	\section{Proposed Models}

To enhance the quality of KGE in order to better perform commonsense knowledge acquisition, we propose a knowledge graph representation learning model in which representations are derived from multitude of information resources. 

At high level, this model can be divided into two main parts. The knowledge-based model captures the inherent structure of the knowledge graph, and the semantic-based model captures the multidimensional aspects of concepts from external semantic resources. Each model is scored through an energy function (i.e., $f(h,r,t) =E(h,r,t)$) and the two models are learned jointly through the following overall energy function:
\begin{equation}
E = E_K + E_S 
\end{equation}

where $E_K$ is the energy function of knowledge-based representations, while $E_S$ is the energy function of semantic-based representations. For each semantic description $E_{S_j}$, semantic and knowledge representations are enforced to be compatible with each other as follow:
\begin{equation}
E_{S_j} = E_{S_jS_j} + E_{S_jK} + E_{KS_j}, \label{compatible_equ}
\end{equation} 
where, 
\begin{equation*}
E_{S_jS_j} = \Vert \mathbf{h_{s_j} + r - t_{s_j}} \Vert,
\end{equation*} 
\begin{equation*}
E_{S_jK} = \Vert \mathbf{h_{s_j} + r - t_k} \Vert,
\end{equation*} 
\begin{equation*}
E_{KS_j} = \Vert \mathbf{h_k + r - t_{s_j}} \Vert.
\end{equation*} 

The overall energy function will project the two types of concept representations into the same vector space while the relation representation is shared and updatee by all energy functions. Numerous KGE models can be used to define $E_K$ (a comprehensive review of these models in~\cite{wang2017knowledge}). \\

\subsection{Knowledge Representation Model}
The knowledge model scores each triple based solely on the internal links, hence capture the local connectivity patterns of the knowledge graph.
In this model, a link between two entities is an operation on their vectors. Some prominent models are: TransE that scores a triple through an energy function which consider a relation as a translation from head to tail entity in the form $f(h,r,t) = \lVert \mathbf{h+r-t} \rVert$, TransR~\cite{lin2015learning} extends TransE such that entities and relations are embedded into distinct entity and relation spaces $\mathbb{R}^{k}$ and $\mathbb{R}^{d}$, respectively.
Another model is structured embedding~\cite{bordes2011learning} that scores a triple via a bilinear score function of form $f(h,r,t) = \mathbf{{v_h}^T M_r v_t}$. 

In this work, we adopt the basic TransE model, thus knowledge model energy is defined as: 
\begin{equation}
E_K = \lVert \mathbf{h_k+r-t_k} \rVert
\end{equation}
where $E_K$ is expected to have a low value for correct triples and high value otherwise.


\subsection{Semantic Representation Model}
\textit{\textbf{1. Textual semantics:}}
Commonsense knowledge base connects concepts, in the form of words and phrases of natural language, with labelled edges.
Knowledge embedding models consider concepts and relations as symbolic elements and recover their structural relatedness and regularities. However, words and phrases as standalone elements carry rich semantic information. Furthermore, their involvement in triples imply their semantic relatedness. Word embeddings, such as Word2vec~\cite{mikolov2013efficient} and GloVe~\cite{pennington2014glove}, capture words generic semantic and syntactic information
from large corpora through optimizing task-independent objective function that is agnostic to their structural connectivity. Inferences involving commonsense concepts can largely benefit from concept semantic embeddings when injected into the knowledge representation learning process. This is particularly true for concepts with few training instances, in which case, degrading the quality of knowledge-model embeddings. Thus, semantic relatedness between two phrases can be modeled as 
\begin{equation*}
\Vert \mathbf{h_t+r-t_t} \Vert
\end{equation*}
where $\mathbf{h_t}$ and $\mathbf{t_t}$ are word embeddings of the two concepts phrases. One way to obtain $\mathbf{h_t}$ and $\mathbf{t_t}$ is by averaging word vectors of $h$ and $t$ . When word and entities embeddings are in different spaces, they are not useful for any computation. To address this, the energy function of the textual semantic model is formulated as in \ref{compatible_equ} to enforce both representations to be compatible:
\begin{equation}
E_{T} = E_{TT} + E_{TK} + E_{KT}
\end{equation}
such that 
\begin{equation}
\begin{aligned}
E_T = \Vert \mathbf{h_{t} + r - t_{t}} \Vert + \Vert \mathbf{h_{t} + r - t_k} \Vert + \\
\Vert \mathbf{h_k + r - t_{t}} \Vert \label{expand_E_T}
\end{aligned}
\end{equation}

The textual semantics model starts by initializing concepts with word embeddings then run the aforementioned energy function to fine-tune them to be consistent with their knowledge embedding counterpart. Word2vec and GloVe are two well known and effective word embeddings that have complimentary strengths over the other. Recently, Speer et al.~\cite{speer2017conceptnet} presented a novel word embedding model called Numberbatch. This model outperformed Word2vec and GloVe in the semantic word similarity task of SemEval 2017\footnote{http://alt.qcri.org/semeval2017/task2}. In fact, it takes Word2vec and GloVe word vectors as input and improve on them by the mean of retrofitting~\cite{faruqui2014retrofitting}, a method to refine existing word embeddings using relation information from external resource. 

Specifically, it is derived from ConceptNet multilingual graph and global word vectors. Given word vectors $\hat{w_i}$ from a word embedding model, retrofitting infers new vectors $w_i$, such that they are close to their original value and close to their neighbors: 
\begin{equation}
\sum_{i}^{m} \Bigg[ \alpha_i \Vert w_i - \hat{w_i}\Vert + \sum_{(i,*,j) \in \pazocal{T}} \beta_{ij} \Vert w_i - w_i \Vert^2 \Bigg]
\end{equation}
where $\alpha$ and $\beta$ values control the relative strengths of associations,$m$ is the size of vocabulary, and $(i,*,j)$ are all concept pairs in the knowledge graph connected by arbitrary relations. Thus, we utilize Numberbatch embeddings as the word embeddings of concepts in our knowledge base.\\

\noindent \textit{\textbf{2. Affective Valence:}}
Recent models for concept-level sentiment analysis associate concepts with values encoding their affective valence information~\cite{camspa,pormer,camact,howint}. These models define a notion of relatedness between concepts according to their semantic and affective valence. AffectiveSpace~\cite{cambria2015affectivespace} is one of such models and allows semantic features associated with concepts to be generalized and, hence, allows concepts to be intuitively clustered according to their semantic and affective relatedness. This vector model lend itself as powerful framework that can be embedded in potentially any cognitive system dealing with real-world semantics. Thus, we inject these affective vectors into knowledge-based representation learning with the aim of discovering potential assertion between concepts based on their affective relatedness. We define the affective semantic energy function $E_A$ as:
\begin{equation}
E_{A} = E_{AA} + E_{AK} + E_{KA}
\end{equation}
where $E_{AA} = \Vert \mathbf{h_a +r - t_a}\Vert$, $\mathbf{h_a}$ is the affective vector produced by AffectiveSpace, and $E_A$ is expanded analogically to \ref{expand_E_T}.\\


\noindent \textit{\textbf{3. Common Knowledge:}}	
``You shall know a word by the company it keeps"~\cite{firth1957synopsis} is a principle that underpinned many text and graph embedding models. For example, Word2vec skip-gram model predicts a word from its context, and graph embedding models such as Deepwalk~\cite{perozzi2014deepwalk}, LINE~\cite{tang2015line} and node2vec~\cite{grover2016node2vec} learn node embeddings based on their first-order or second-order neighborhood. 
In the same vein, but for commonsense concepts, Chen and de Melo~\cite{chen2015semantic} suggested using concept definitions and lists as focused contexts for concept embeddings. Inspired by this work, we propose new semantic context definition that have a potential to provide a boast in concepts embeddings expressiveness.

Since concepts are high level abstractions and given the implicit nature of their mentions, their diverse meanings might be difficult to retrieve from text. One way to recover some of these meanings is through examining instances connected with concepts via hyponym-hypernym relations. These instances carry sub-meanings of their more general superordinates, thus, theoretically should have similar embeddings, but more importantly should carry more focused semantic inferences. 

In our model, we aim to recover as much as possible of instances categorized under each concept and integrate their embeddings into our knowledge model. That is, for each concept $c \in \pazocal{C}$, we retrieve a list of instances $I_c = \{I_{c,1}, I_{c,2},.., I_{c,n}\}$, where $I_{c,j}$ is the $j^{th}$ instance of concept $c$ and $n$ is the total number of instances of concept $c$. These instances are then used to construct \textit{common-knowledge} embedding $\mathbf{c_{c}}$. 
Assuming each instance $I_{c,i}$ has embedding $\mathbf{I_{c,i}}$, the \textit{common-knowledge} embedding of concept $c$ is defined as:



\begin{equation}
\mathbf{c_{c}} = \frac{1}{n} \sum_{I_{c,i} \in I_{c}} \mathbf{I_{c,i}}
\end{equation}

The final semantic energy $E_C$ function for this external resources is then:
\begin{equation}
E_{C} = E_{CC} + E_{CK} + E_{KC}
\end{equation}
and it is expanded analogically to \ref{expand_E_T}.

\subsection{Training}
To obtain entities and relations embedding, the model aims to maximize the following margin-based objective function that discriminate between correct triples and incorrect triples: 
\begin{equation}
\begin{aligned}
\pazocal{L}= \sum_{(h,r,t) \in \pazocal{T}} \sum_{(h^\prime,r^\prime,t^\prime) \in \pazocal{T^\prime}} \text{max}(0, \gamma + E(h,r,t) - \\
E(h^\prime, r^\prime, t^\prime) )
\end{aligned}
\end{equation}

where $\text{max}(.,.)$ return the maximum of two inputs, $\gamma$ is the margin hyper-parameter, $ \pazocal{T}$ denote golden triples $(h,r,t)$, and $ \pazocal{T}^\prime$ denote corrupted triples: $\lbrace \lbrace h^\prime,r,t) | h^\prime \in \pazocal{C},(h^\prime,r,t) \notin \pazocal{G} \rbrace 
\cup 
\lbrace (h,r,t^\prime) | t^\prime \in \pazocal{C},(h,r,t^\prime) \notin \pazocal{G} \rbrace
\cup 
\lbrace (h,r^\prime,t) | r^\prime \in \pazocal{R},(h,r^\prime,t) \notin \pazocal{G} \rbrace \rbrace$. 
We adopt stochastic gradient descent (SGD) to minimize the above loss function. 

We train our model with two settings. At first, we initialize concepts with the pre-compiled semantic representations described above.
In \textit{Fixed} setting, during training we fix concepts' auxiliary semantic representations and only update knowledge-based concept and relation representations. In \textit{variable} setting, we update all representations simultaneously.



\section{Experimental Results}

We empirically evaluate our models with regards to two tasks: knowledge base completion and triple classification. We experiment with each of the individual semantic models separately then combine them together and compare with the knowledge model as a baseline.

\subsection{Datasets and Experiment Settings}
\textbf{Knowledge Base. } We derive the dataset from the English part of ConceptNet which contains around 1,803,873 concepts, 38 relations, and 28 million triples. In particular, we keep concepts that have counterparts in the semantic resources we mentioned above. 
We end up with a knowledge base of 30773 concepts, 38 relations, and 366202 triples, lets call it here CN30K for simplicity. These triples were then divided into training, validation, and test sets. 

To make these three sets balanced with enough training examples for each relation type, we first count triples of each relation, then we divide them with 60\%, 20\%, and 20\% ratios accounting for $240246$, $63992$, and $61964$ triples for train, validation, and test respectively.
The resulting knowledge base is highly skewed with majority of triples are connect by generic relations, e.g., $80\%$ of triples are connected via {\tt RelatedTo, Synonym}, and {\tt IsA} relations, while relations such as { \tt NotHasProperty, CreatedBy, InstanceOf, ReceivesAction, DefinedAs, LocatedNear, MannerOf, NotCapableOf}, and {\tt SymbolOf} made up around $1\%$ of triples.\\

\textbf{Affective Valence. }
We associate each concept in CN30K with a vector encoding its affective valence. We use AffectiveSpace, a model built by means of random projection to reduce the dimensionality of affective commonsense knowledge. Specifically, the random projection was applied on the matrix representation of AffectNet, a commonsense knowledge base built upon ConceptNet and WordNet-Affect~\cite{strapparava2004wordnet}, an extension of WordNet Domains, including a subset of synsets suitable to represent affective concepts correlated with affective words.\\

\textbf{Common knowledge Instances. }
A straightforward way to obtain concept instances is by inquiring other knowledge bases such as DBpedia~\cite{lehmann2015dbpedia} or WebChild~\cite{tandon2017webchild} for entities connected to concepts via IsA relation, i.e., (\textit{Entity}, { \tt IsA}, \textit{Concept}). Probase is a probabilistic taxonomy of common knowledge organized as a hierarchy of hyponym-hypernym relations. It consists of 5,401,933 unique concepts and 12,551,613 unique instances harnessed from 1.68 billion web pages. We consider this knowledge base as a source to obtain the concepts subordinates. Probase concepts are expressed as raw textual phrases. We thus start by running concept parser\footnote{http://stanfordnlp.github.io/CoreNLP} over concepts' phrasal expressions to generate their term expressions, e.g., phrasal concept: ``\textit{established high fashion brand}" generates concept's terms \textit{high\_brand, fashion\_brand,} and \textit{establish\_brand}. Each Probase concepts will have one or more term expressions. Then, we match concepts in CN30K with concept terms of Probase. Afterwards, for each matched Probase concept, we retrieve a list of instances $I_c = \{I_{c,1}, I_{c,2},.., I_{c,n}\}$ and associate them with CN30K counterpart.

\begin{table*}[tb]
	
	\begin{center}		
		\renewcommand{\arraystretch}{.9} 
		\small 
		{
			\begin{tabular}{p{.75in} |C{.3in} C{.3in}|C{.45in} C{.45in}||C{.3in} C{.3in}|C{.45in} C{.45in}}
				\toprule
				&\multicolumn{4}{c||}{\bfseries Fixed} & \multicolumn{4}{c}{\bfseries Variable} \tabularnewline
				
				\hline
				
				\multirow{2}{*}{\bfseries Model} & 
				
				\multicolumn{2}{c|}{\bfseries Mean Rank} & \multicolumn{2}{c||}{\bfseries Hits@10(\%)}& \multicolumn{2}{c|}{\bfseries Mean Rank} & \multicolumn{2}{c}{\bfseries Hits@10(\%)} \tabularnewline
				& Raw & Filter & Raw & Filter 	& Raw & Filter & Raw & Filter \tabularnewline
				\hline	
				\tabularnewline			
				\textsc{TransR} &3648 & 3628 & 2.99\% & 3.20\% & 3648 & 3628 & 2.99\% & 3.20\% \tabularnewline	
				\textsc{TransE} & 2477 & 2453 & 19.77\% & 24.29\% & 2477 & 2453 & 19.77\% & 24.29\% \tabularnewline				
				\textsc{TransE+TXT} & \textbf{1059} & \textbf{1039} & \textbf{22.97}\% & \textbf{26.59}\% & 1259 & 1235 &21.18\% &\textbf{ 26.49}\% \tabularnewline
				\textsc{TransE+AFF} & 3749 & 3728 & 10.36\% & 11.08\% & 1502 & 1478 & 20.56\% & 25.48\% \tabularnewline	
				\textsc{TransE+CK} & 3113 & 3093 & 7.39\% & 7.95\% & 1386 & 1362 & 20.18\% & 24.83\% \tabularnewline
				\textsc{TransE+ALL} & 1654 & 1634 & 16.88\% & 18.78\% & \textbf{1089} & \textbf{1065} & \textbf{21.29}\% & 26.37\% \tabularnewline
				\bottomrule				
			\end{tabular}
		}
		\caption{\label{table:concept_prediction} Evaluation results on concept prediction}
	\end{center}
\end{table*}


\begin{table*}[tb]
	\begin{center}
		\renewcommand{\arraystretch}{.9}
		\small 
		{
			\begin{tabular}{p{.75in} |C{.3in} C{.3in}|C{.45in} C{.45in}||C{.3in} C{.3in}|C{.45in} C{.45in}}
				\toprule
				&\multicolumn{4}{c||}{\bfseries Fixed} & \multicolumn{4}{c}{\bfseries Variable} \tabularnewline
				\hline
				\multirow{2}{*}{\bfseries Model} & 
				
				\multicolumn{2}{c|}{\bfseries Mean Rank} & \multicolumn{2}{c||}{\bfseries Hits@10(\%)}& \multicolumn{2}{c|}{\bfseries Mean Rank} & \multicolumn{2}{c}{\bfseries Hits@10(\%)} \tabularnewline
				& Raw & Filter & Raw & Filter 	& Raw & Filter & Raw & Filter \tabularnewline
				\hline	
				\tabularnewline
				\textsc{TransR} & 25.05 & 24.90 & 18.22\% & 18.33\% & 25.05 & 24.90 & 18.22\% & 18.33\% \tabularnewline					
				\textsc{TransE} & 11.86 & 11.73 & 30.58\% & 31.24\% & 11.86 & 11.73 & 30.58\% & 31.24\% \tabularnewline
				\textsc{TransE+TXT} & 10.53 & 10.4 & 35.33\% & 36.26\% & 10.08 & 9.95 & 43.68\% & 44.85\% \tabularnewline	
				\textsc{TransE+AFF} & 3.899 & 3.784 & \textbf{95.57}\% & \textbf{95.74}\% & 4.303 & 4.179 & 92.02\% & 92.44\% \tabularnewline
				\textsc{TransE+CK} & 8.629 & 8.488 & 66.16\% & 66.98\% & \textbf{2.446} & \textbf{2.333} & \textbf{94.62}\% &\textbf{ 94.91}\% \tabularnewline	
				\textsc{TransE+ALL} & \textbf{3.625} &\textbf{3.51} & 93.2\% & 93.57\% & 5.093 & 4.969 & 90.69\% & 91.2\%\tabularnewline
				\bottomrule	
			\end{tabular}
		}
		\caption{\label{table:relation_prediction} Evaluation results on relation prediction}
	\end{center}
	
\end{table*}

\subsection{Knowledge Graph Completion} 
Used in~\cite{bordes2013translating}, the task of knowledge graph completion aims to complete a triple $(h,r,t)$ when one of $h, t, r$ is missing.
Instead of only giving one best answer, the score function $f(h, r, t)$ ranks a set of candidate concepts and relations from the knowledge graph. The knowledge graph completion task has two sub-tasks: concept prediction and relation prediction. The result of each sub-task is reported separately. \\ 


\noindent \textbf{Evaluation Protocol. } For each test triple $(h,r,t)$, we replace the head/tail concepts by all concepts in the knowledge graph then ranked them in ascending order of dissimilarity scores. The same procedure is performed for relation predication. We use two measures as our evaluation metrics: (1) mean rank of correct concepts; (2) proportion of valid concepts ranked in top 10. A good predictor should achieve lower mean rank and higher Hits$@$10. This basic setting of the evaluation is called ``Raw" setting, called so because all concepts in the knowledge graph are evaluated and ranked. However a number of corrupted triples may end up being valid ones from training or validation sets, and the model well be penalized for ranking corrupted triple higher than test triple. To eliminate this issue, in the ``Filter" setting, corrupted triples that exist in either the train, validation and test datasets are filtered out. \\

\noindent \textbf{Implementation.} To train our model, we use learning rate $\alpha$ for SGD among $\{0.001, 0.005, 0.01\}$, the margin $\gamma$ among $\{0.25, 0.5, 1, 2\}$, and the embedding dimension $n$ among $\{50,80,100\}$. We further used a fixed batch size of $5000$.
The optimal parameters are determined by the validation set. The optimal configurations are: $\alpha = 0.01, \textrm{ and } \gamma = 1, n = 100$.\\

%

\noindent \textbf{Results.} We consider TransE and TransR as baseline models and we compare their performance with that of individual semantic models and with the composition of knowledge model and all semantic models, denoted TransE+ALL. The results of concept predication and relation predication in both \textit{Fixed} and \textit{Variable} settings are shown in Table~\ref{table:concept_prediction} and Table~\ref{table:relation_prediction}. 
TransE and TransR are run under one \textit{Variable} setting, since there are no auxiliary information. We notice that in concept and relation prediction, TransE perform better than TransR.

Under \textit{Fixed} setting, we notice that the textual semantic model TransE+TXT deliver the best performance in concept predication while at the same time show improvements over the baseline in relation prediction. The other models, however, show extreme discrepancy in performance in both tasks. For example, the TransE+AFF and the TransE+CK models have rather poor results in concept predication while delivering remarkable improvements in relation prediction. 

These are understandable results, since the textual semantic representations were optimized to encode not only words semantics, but also words structural connectivity in a relational knowledge, therefore, they transfer some of concepts relational similarities to relations representations, hence the stability in performance. \comment{therefore they where able to serve the semantic similarity in concept prediction and relational similarity in relation prediction.} However, in case of affective valence and the common knowledge semantic models, their representations do not encode any structural information, therefore the vectors learned by TransE+AFF and TransE+CK target relation prediction exclusively, irrespective to concepts structural connectivity.
Under \textit{Variable} settings, however, TransE+AFF and TransE+CK show better generalization capability with continuing to show prominent results for relation prediction, but this time without deteriorating their effectiveness in concept prediction. 

In fact, they show comparable results with TransE baseline in concept prediction, while TransE+TXT still show the same consistent behaviour with improvements over both tasks and showing the best performance in concept prediction. Notably, TransE+CK has the highest improvement over all other models in relation prediction, confirming thereby that gaining insight into concept meanings (from its instance) help recover structural regularities that are more evident in factual knowledge.	

Finally, we remark that TransE+ALL get affected by the least performing models in all settings, however, combining highest performing models only is believed to perform better that any.

\subsection{Triple Classification}
Triples classification aims to judge whether a given triple (h, r, t) is correct or not, which is a binary classification task.

\noindent \textbf{Evaluation Protocol. } Naturally, a classification task needs samples with positive and negative labels in order to learn a discriminative classification model. Thus, we construct a negative samples for our training set as follows: for each golden triple we generate three negative triple by randomly switching one of $h,r,t$ at a time with $h^\prime,r^\prime,t^\prime$, such that $\{h^\prime,r^\prime,t^\prime\} \in \pazocal{C}$, and $((h^\prime,r,t),(h,r,t^\prime),(h,r^\prime,t)) \notin \pazocal{G}$. 

The classification decision rule is as follows: for a given triple $(h, r, t)$, if its score is less than relation-specific threshold $\delta_r$, it will be classified as positive, otherwise it considers as negative. $\delta_r$ is obtained by maximizing the classification accuracies on the valid set. 


\noindent \textbf{Implementation. } We apply the same parameter settings as in previous task. 

\noindent \textbf{Result. } We measure our model'��s ability to discriminate between golden and corrupted triples. From Table~\ref{table:classification_accurecy}, we can see that TransR delivers a better performance than TransE and some other enhanced models. 

We can also see that in both \textit{Fixed} and \textit{Variable} settings, the TransE+CK semantic model have the highest classification accuracy with comparable result to TransR. This interesting observation can be further investigated in future work to examine the effectiveness of similar semantic enhancements on TransR model.

 We also observe that TransE+AFF have surprisingly better performance than TransE+TXT, and in \textit{Variable} scenario, outperform the baseline. These results are strong indication of effectiveness of semantic models in equipping concepts with discriminative features. Hence resolving part of existing ambiguity and commonsense reasoning in an effective manner.

\begin{table}[t] 
	\begin{center}
		\renewcommand{\arraystretch}{1}
		\small 
		{
			
			\begin{tabular}{C{1in} |C{.75in} C{.75in}}
				\toprule
				
				Model&\multicolumn{2}{c}{\bfseries Accuracy} \tabularnewline
				
				& Fixed & Variable\tabularnewline
				\hline
				\textsc{TransR} & 91.04	 & 91.04 \tabularnewline
							
				\textsc{TransE} & 88.73	 & 88.73 \tabularnewline
				\textsc{TransE+TXT} & 83.66& 88.75 \tabularnewline	
				\textsc{TransE+AFF} & 87.85 &90.41 \tabularnewline
				\textsc{TransE+CK} & \textbf{92.94} &\textbf{91.72} \tabularnewline	
				\textsc{TransE+ALL} & 90.23 & 89.59 \tabularnewline
				\bottomrule	
			\end{tabular}
		}
		\caption{\label{table:classification_accurecy} Evaluation results on Triple classification}
	\end{center}
	
\end{table}


\section{Conclusion}
We investigate enhanced KGE models aiming to improve automatic commonsense knowledge acquisition. In particular, we consider models that perform joint representation learning from structural and semantic resources. We derive a set of semantically salient resources that cover structural, semantic, affective and taxonomical aspects of concepts. We run a joint model to bring knowledge graph structural representation and the auxiliary semantic resources representations into the same vector space. 

Empirical results show that semantic information is indeed effective and has the potential to further improve commonsense knowledge acquisition. As future work, we plan to investigate learning knowledge and semantic representations simultaneously rather than following a two-step model. We would also like to introduce a modification to the common knowledge model such that each concept would have multiple semantic representations based on the number of categories in its instances list.

\bibliographystyle{IEEEtran}
\bibliography{references}

\end{document}